\documentclass[11pt,a4paper]{article}
\usepackage{times,latexsym}
\usepackage{url}
\usepackage[T1]{fontenc}
\usepackage{acl} 

\usepackage{microtype}
\usepackage{graphicx}
\usepackage{amsmath}
\usepackage{amssymb}
\usepackage{booktabs}
\usepackage{tabularx}
\usepackage{adjustbox}
\usepackage{enumitem}
\usepackage{float}

\newcommand{\KL}{\mathrm{KL}}
\newcommand{\TV}{\mathrm{TV}}
\newcommand{\flips}{\mathrm{flip}}
\newcommand{\rhoc}{\rho_c(0)}

\newenvironment{fittable}[1][\columnwidth]%
  {\begin{adjustbox}{max width=#1}}{\end{adjustbox}}

\newcommand{\figorbox}[1]{%
  \IfFileExists{#1}{\includegraphics[width=\columnwidth]{#1}}{%
    \fbox{\parbox[c][3.2cm][c]{0.92\columnwidth}{\centering\small\texttt{\detokenize{#1}}\\(figure pending)}}}}

\title{How Divergence Becomes Decision Flips in \\ Compressed Language Models}

\author{Beatriz Almeida Felicio \\
  Instituto de Informática (INF) \\
  Universidade Federal de Goiás (UFG) \\
  \texttt{beatrizfelicio@discente.ufg.br}}

\date{}

\begin{document}
\maketitle

\begin{abstract}
Compression reports summarize how far a compressed language model moved from the dense one, usually by a KL divergence; a deployment that relies on the dense model's outputs needs to know how many of its decisions changed.
We show that total variation, not KL, answers this directly.
Across 802 compressed and perturbed copies of 19 open models on five corpora and nine mechanically unrelated perturbation families, the rate at which the arg-max token changes (the \emph{flip rate}) tracks total variation at a ratio with median $1.05$, with no fitted constant.
KL converts into flips only through its square root and a factor that varies fourfold across models and corpora, because KL averages over tokens before the root is taken; first-order statistics averaged per token, such as Hellinger distance, avoid this, but reports rarely give them.
As a result, of two compressors reported on different models and corpora whose flip rates differ by at least $10\%$, KL assigns the smaller divergence to the one that changes more decisions in $11\%$ of cases, total variation in $1\%$.
Two pre-registered tests mark the limits: on a held-out code corpus the ratio held for all eight models while three predictions about KL each failed for half of them or more, and on three new models with real kernels it stayed in its band for 37 of 38 checkpoints but fell below one on code for two models.
In vLLM speculative decoding, total variation measured under teacher forcing predicts greedy draft acceptance with a mean relative error of $1.1$--$2.4\%$, without the task-specific calibration that KL needs.
\end{abstract}

\begin{figure*}[t]
  \centering
  \includegraphics[width=\textwidth]{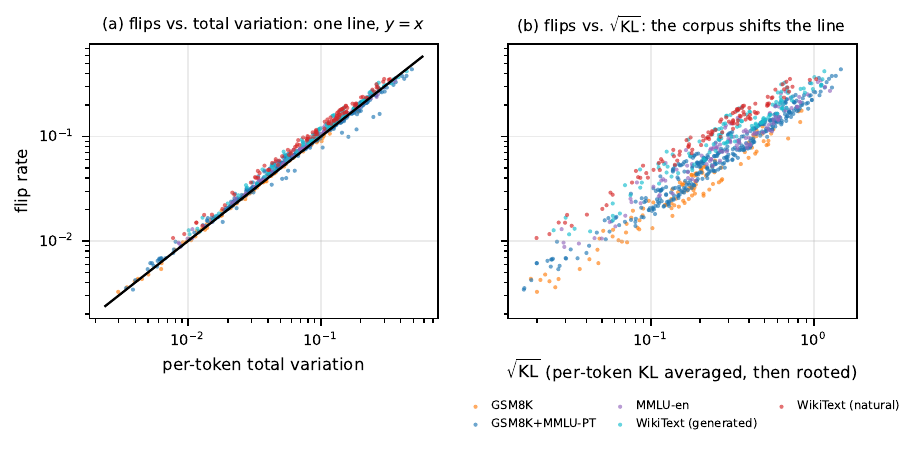}
  \caption{The same 802 compressed and perturbed configurations, measured two ways. \emph{(a)} Against total variation, the flip rate lies close to the identity line with no fitted constant. \emph{(b)} Against $\sqrt{\KL}$, the relation is a power law whose level varies by corpus: on GSM8K, where a few tokens carry the divergence, the same KL means fewer changed decisions than on natural text.}
  \label{fig:teaser}
\end{figure*}

\section{Introduction}
\label{sec:intro}

Post-training compression is the standard way to reduce the memory and bandwidth cost of Transformer language models \citep{vaswani2017attention,gholami2021survey}.
Weight quantization \citep{dettmers2022llmint8,frantar2023gptq,lin2024awq,xiao2023smoothquant}, pruning \citep{frantar2023sparsegpt,sun2024wanda,ma2023llmpruner}, KV-cache quantization \citep{liu2024kivi,hooper2024kvquant} and layer removal \citep{men2024shortgpt,gromov2024unreasonable} are typically assessed end to end: apply the method, run benchmarks, report the change in accuracy \citep{li2024evaluating,jaiswal2024compressing}.
End-to-end accuracy mixes (i) how far the method moves the model's predictive distribution, (ii) how that movement turns into changed token-level decisions, and (iii) how changed decisions propagate through generation into changed answers.
For several consumers only stage (ii) matters: systems that need exact agreement with the dense model, such as cached or deduplicated outputs and regression tests across model versions, depend on how many token decisions change.
Speculative decoding is the clearest case: it accepts a draft token only when it agrees with the target; under sampling, the acceptance probability is exactly one minus the total variation between the two next-token distributions \citep{leviathan2023speculative}, and under greedy decoding a token is accepted exactly when the two arg-max decisions agree.
The sampling identity is known; whether a teacher-forced statistic predicts greedy agreement has, to our knowledge, not been tested, and \S\ref{sec:spec} tests it.
Stage (iii) is outside this paper, and it bounds everything else: fidelity to the dense model and task accuracy come apart (\S\ref{sec:discussion}), so the statistics studied here describe decisions, not benchmark quality.

A line of work already argues that accuracy hides compression damage and proposes distributional statistics instead: \citet{dutta2024accuracy} report \emph{flips} alongside KL divergence, \citet{qamar2026accuracy} evaluates quantized models by total variation, and \citet{kishino2026scale} calibrates the scale of KL.
What prior work leaves open is the quantitative relation between the statistics themselves: which divergence converts into changed decisions, through what functional form, whether one conversion serves unrelated compression methods, and why it depends on the corpus.
Theory does not settle it: total variation bounds the flip rate (Lemma~1, \S\ref{sec:kappa}), but the bound allows several times more flips than compressors produce.
The practical question is how to read a reported number.
Papers, leaderboards and tools report KL to the dense model and compare it across methods and evaluation sets; someone choosing between two quantized releases, or deciding whether a lower KL is worth a slower kernel, needs to know how many decisions each number implies and whether two numbers measured on different text can be compared at all.
Counting flips answers this only for the model and corpus at hand; the divergences that exist in reports, usually KL averaged over tokens, are what must be read.

We therefore isolate stage (ii).
Under teacher forcing we measure, at the same positions and on the same prefixes, the divergence between a dense model and a perturbed copy and the rate at which the arg-max token changes, and ask three questions:
\textbf{RQ1}: how does the flip rate scale with divergence, and does one conversion serve all perturbation families?
\textbf{RQ2}: what sets the conversion, and why does it depend on the corpus?
\textbf{RQ3}: does the answer predict greedy speculative decoding, the consumer that prices decisions directly?

Our contributions are:
\begin{enumerate}[itemsep=1pt,topsep=2pt,leftmargin=*]
\item \textbf{Total variation reads as decisions.} Flips per unit of total variation have median $1.05$ ($95\%$ of configurations within $0.89$--$1.29$) across nine families, with no fitted constant; an isotropic small-noise limit that depends on the dense distributions alone accounts for the value near one. The ratio held in two pre-registered tests, except on code for two new models (\S\ref{sec:kappa}, \S\ref{sec:holdout}).
\item \textbf{KL reads as decisions only within a corpus.} Within one reference (a model on a corpus), nine families share one square-root curve; across models and corpora its coefficient moves fourfold ($1.6$--$2.6\times$ across corpora of one model), so KL misorders compressors compared across reports eight times as often as total variation, and on a held-out code corpus the shared family curve failed its registered test (\S\ref{sec:law}--\S\ref{sec:holdout}).
\item \textbf{Why: averaging.} An exact decomposition attributes most of KL's corpus dependence to its averaging over tokens (a Jensen factor); first-order statistics averaged per token, such as Hellinger distance, share total variation's stability, but reports give aggregate KL (\S\ref{sec:kappa}).
\item \textbf{Greedy speculative decoding.} Total variation measured under teacher forcing predicts a draft's greedy agreement without task-specific calibration, where KL must be calibrated on the task and still errs two to three and a half times more; in vLLM it predicts measured acceptance within $1.1$--$2.4\%$ (\S\ref{sec:spec}).
\end{enumerate}

\section{Related Work}
\label{sec:related}

\paragraph{Compression and its evaluation.}
Post-training quantization \citep{frantar2022obc,frantar2023gptq,lin2024awq,yao2022zeroquant,chee2023quip,ashkboos2024quarot,egiazarian2024aqlm,dettmers2023spqr,shao2024omniquant,kim2024squeezellm} and pruning \citep{han2015learning,frantar2023sparsegpt,sun2024wanda,ashkboos2024slicegpt,men2024shortgpt} are usually compared by accuracy or perplexity.
Several of these methods already minimize the second-order expansion of the loss that we use descriptively to explain the exponent \citep{nagel2020adaround,li2021brecq,dong2019hawq}.
Accuracy and perplexity are known to hide differences between methods, bit widths and tasks \citep{jaiswal2024compressing,li2024evaluating,kurtic2024bf16,huang2024llama3quant,dettmers2023kbit}, including instance-level disagreement at unchanged accuracy \citep{rababah2026illusion}.

\paragraph{Divergence and flips as evaluation statistics.}
\citet{dutta2024accuracy} propose flips and KL as complementary metrics for compressed models, and \citet{xu2021loyalty} earlier defined \emph{label loyalty} (a flip rate) and \emph{probability loyalty} (a distributional distance) for compressed BERT models; \citet{hooker2019compressed} ask which examples compression forgets, and KL and flips are also reported side by side for learned mixed precision \citep{ogilvy2026vbq}.
More recent work evaluates quantized models by total variation and Jensen--Shannon divergence \citep{qamar2026accuracy}, sets an empirical scale for KL across settings \citep{kishino2026scale}, and finds that KL tracks the volume of disagreement rather than its direction \citep{nikolic2026displacement}.
These works treat divergence as the endpoint; we take it as the input and ask what it implies for decisions, how the answer depends on the divergence chosen, and why it varies with the corpus.
The square-root form follows from standard information geometry \citep{amari1998natural,martens2020fisher}.
Table~\ref{tab:related} lists, point by point, what the closest studies measure: \citet{qamar2026accuracy} already evaluate by total variation and Jensen--Shannon divergence, but none of these studies fits a functional relation between a divergence and the flip rate, none accounts for why that relation depends on the corpus, and none tests it on a held-out domain.

\begin{table}[t]
  \centering
  \scriptsize
  \setlength{\tabcolsep}{2pt}
  \renewcommand{\arraystretch}{1.05}
  \begin{tabularx}{\columnwidth}{@{}>{\raggedright\arraybackslash}p{0.19\columnwidth}>{\raggedright\arraybackslash}X>{\raggedright\arraybackslash}p{0.2\columnwidth}>{\raggedright\arraybackslash}p{0.22\columnwidth}>{\raggedright\arraybackslash}p{0.12\columnwidth}@{}}
    \toprule
    study & statistics & methods & divergence $\to$ flips & domain transfer \\
    \midrule
    \citet{dutta2024accuracy} & flips, KL & 6 quant.\ schemes, layer drop, Wanda, SliceGPT & correlation; margin account of flips & no \\
    \citet{qamar2026accuracy} & TV, JS, accuracy & llama.cpp formats & only evaluates endpoint, no map to flips & no \\
    \citet{kishino2026scale} & KL & quantization among other settings & fits empirical scale, no mechanistic breakdown & no \\
    \citet{nikolic2026displacement} & KL variants, top-1 agreement & existing quantized releases & reports task correlation, no functional form & no \\
    \midrule
    this work & flips, KL, TV, JS, Hellinger & 9 families, MoE, QAT & ratio $\approx1$ (TV), power law (KL), bound, decomposition & code, registered \\
    \bottomrule
  \end{tabularx}
  \caption{What the closest studies of divergence and flips in compressed language models measure. \emph{Divergence $\to$ flips}: whether a relation between a divergence and the flip rate is fitted and explained.}
  \label{tab:related}
\end{table}

\paragraph{Churn, negative flips and margins.}
Disagreement induced by retraining is studied as \emph{churn} \citep{fard2016launch,bhojanapalli2021reproducibility} and as \emph{negative flips} \citep{yan2021pct}; we instead control the perturbation of a fixed model.
Margins govern robustness to input noise in randomized smoothing \citep{cohen2019certified}; ours is the parameter-space counterpart, where the density of margins near zero sets how many decisions change.

\paragraph{Speculative decoding.}
Speculative decoding \citep{leviathan2023speculative,chen2023speculative,stern2018blockwise,cai2024medusa,li2024eagle,miao2023specinfer} accepts a draft token when it agrees with the target, and self-speculative variants draft with a compressed copy of the target \citep{zhang2024draft,tiwari2025quantspec,zhao2025qspec}; for sampling, acceptance equals one minus the total variation between the two distributions \citep{leviathan2023speculative}, KL has been argued to be the wrong training objective for drafts \citep{zhou2024distillspec,hu2025adaspec}, and total variation has been used directly as a training loss for generation and for drafts \citep{ji2023tv,goel2024direct}; we study it as an evaluation statistic.

\section{Setup and Definitions}
\label{sec:setup}

\paragraph{Models, perturbations, corpora.}
We use 19 checkpoints from six families: Gemma-3 (1B, 4B, 12B, 27B) \citep{gemma3}; Qwen3 (4B, 8B, 14B) and Qwen2.5 (7B, 72B) \citep{qwen3,qwen25}; Mistral-7B and Mistral-Small-24B \citep{jiang2023mistral}; Phi-3.5-mini and phi-4 \citep{abdin2024phi3}; OLMo-2 (1B, 7B, 13B), with base and instruction-tuned versions of the 1B and 7B models \citep{olmo2}; and Llama-3-8B-Instruct \citep{grattafiori2024llama3}, loaded from a public mirror whose weight files have the same SHA-256 hashes as the official release.
Perturbations cover nine families: round-to-nearest quantization (RTN, 3--8 bits, group 128), GPTQ \citep{frantar2023gptq}, AWQ \citep{lin2024awq}, SparseGPT \citep{frantar2023sparsegpt}, Wanda \citep{sun2024wanda}, magnitude pruning \citep{han2015learning}, KV-cache quantization \citep{liu2024kivi}, layer removal \citep{men2024shortgpt} and isotropic Gaussian weight noise, plus two controls (single-layer RTN, and four third-party GPTQ/AWQ checkpoints that check our compression code).
The four calibrated methods use 128 windows of 512 tokens from the WikiText-2 training split; because two of our corpora are drawn from that split, \S\ref{sec:law} repeats them with C4 calibration.
The five reference corpora are \textsc{mix} (GSM8K \citep{cobbe2021gsm8k} and MMLU machine-translated to Portuguese, 512 prompts), \textsc{gsm8k}, \textsc{mmlu-en} \citep{hendrycks2021mmlu}, \textsc{wikitext} (model continuations of WikiText-2 excerpts \citep{merity2017pointer}) and \textsc{wikitext-nat}, in which the scored sequence is the actual WikiText continuation; the last is needed for base models, which loop under greedy decoding, and scores every model on identical tokens.
\textsc{mix} was the development corpus of the study; its Portuguese half enters the analyses only as a control for translation (\S\ref{sec:kappa}), not as a second language axis, which would need native corpora.
The design is unbalanced: \textsc{mix} holds $45\%$ of the configurations and RTN with its single-layer variant $35\%$; App.~\ref{app:robust} repeats the family comparison without either, and App.~\ref{app:coverage} lists which references enter each analysis.
All models run in bf16 on H100 GPUs. Compression is simulated (real kernels are tested in \S\ref{sec:holdout}; App.~\ref{app:coverage}), so methods whose payoff is latency are compared by their decisions only.

\paragraph{References and teacher forcing.}
For each model $M$ and corpus $C$, a \emph{reference} is built by decoding the dense model greedily on the prompts of $C$ (or taking the natural text), giving sequences $y_{1:T}$; a (model, corpus) pair is our unit of analysis, 54 in total.
Every perturbed copy is scored by teacher forcing on these sequences, so at position $t$ both models condition on the same prefix $y_{<t}$.
With $p_t$ and $q_t$ the next-token distributions of the dense and perturbed model,
\begin{align}
\KL &= \tfrac{1}{T}\textstyle\sum_{t} \sum_{v} p_t(v)\log\frac{p_t(v)}{q_t(v)}, \label{eq:kl}\\
\TV &= \tfrac{1}{T}\textstyle\sum_{t} \tfrac12\sum_{v} |p_t(v)-q_t(v)|, \label{eq:tv}\\
\flips &= \tfrac{1}{T}\textstyle\sum_{t} \mathbf{1}\!\left[\arg\max p_t \neq \arg\max q_t\right]. \label{eq:flip}
\end{align}
We use the forward direction of KL, with the expectation under the dense model, because the question is how the dense model's own decisions are disturbed.
Both divergences are computed over the top-64 tokens of $p_t$ plus an aggregated tail; over the full vocabulary they are a median $1.003$ (KL) and $1.001$ (TV) times larger (App.~\ref{app:coverage}).

\paragraph{The conversion factor.}
We call the ratio of flips to $\sqrt{\KL}$ the \emph{conversion factor} between divergence and decisions: how many arg-max changes a unit of $\sqrt{\KL}$ produces.
We define $r_i = \flips_i/\sqrt{\KL_i}$ as the per-configuration ratio, used when comparing dispersions across configurations.
The fitted coefficient of a reference is
\begin{equation}
\kappa = \operatorname*{median}_{i \in \mathcal{W}} r_i,
\label{eq:kappa}
\end{equation}
over the configurations of the reference in the window $\mathcal{W} = \{i: 10^{-3} < \KL_i < 0.05\}$,
a median of 4 configurations per reference (range 1--39; App.~\ref{app:coverage}).

\paragraph{Statistical protocol.}
Configurations within a reference share the corpus, the model and its margins, so the 54 references, not the 802 configurations, are the independent units: we fit exponents per reference, treat pooled $R^2$ as descriptive, and compute every interval across references or models, never across configurations nested in them.
In-sample fits are weak evidence here: with a dozen configurations over two or three decades of KL, a per-reference $R^2$ near one is nearly automatic for any smooth monotone relation, so our claims rest on held-out families, held-out configurations and a held-out corpus instead.
Logits in bf16 fall on a grid of $0.0625$--$0.25$ nats, so $0.15$--$6.21\%$ of greedy steps (median $1.16\%$) have an exact top-1/top-2 tie; ties dominate flips at the smallest perturbations, so every analysis of margins uses 42 references rescored with an fp32 output head.
Apart from the two tests of \S\ref{sec:holdout}, no analysis was pre-registered; the design grew as results raised questions.

\section{A Square-Root Regularity Shared Across Families}
\label{sec:law}

In brief: within a reference, flips grow almost exactly as the square root of KL, for a reason that is a matter of orders, and nine mechanically unrelated families fall on the same curve.

\begin{figure}[t]
  \centering
  \figorbox{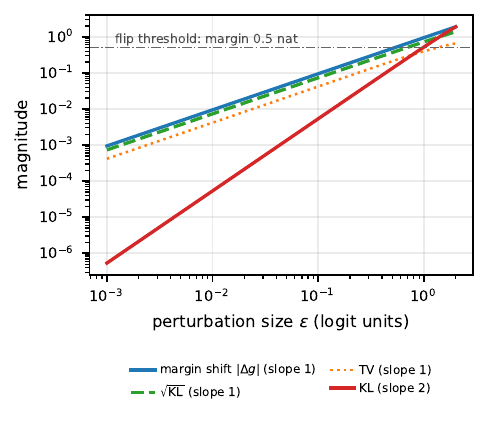}
  \caption{Why a square root, on one illustrative next-token distribution (top-2 margin $0.5$ nat) under a logit shift $\epsilon u$ with fixed random direction $u$: the margin shift, $\sqrt{\KL}$ and total variation grow linearly in $\epsilon$, KL quadratically. A flip needs a shift toward the boundary larger than the margin.}
  \label{fig:orders}
\end{figure}

\begin{figure}[t]
  \centering
  \figorbox{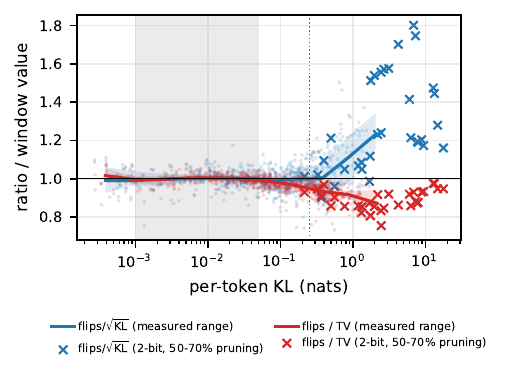}
  \caption{Where the square-root relation bends. Each configuration's $\flips/\sqrt{\KL}$ (blue) and $\flips/\TV$ (red), divided by the reference's value in the $\kappa$ window (grey band); lines are binned medians with interquartile bands, crosses are 32 configurations outside the measured range that enter no fit. The dotted line marks $0.25$ nats.}
  \label{fig:regime}
\end{figure}

\paragraph{Why a square root.}
The intuition is a matter of orders. A small perturbation moves every logit by a small amount $\epsilon$. The gap between the two leading tokens, which decides whether the arg-max changes, moves in proportion to $\epsilon$. KL, which compares whole distributions and is smallest when nothing changes, can only grow as $\epsilon^2$. Measuring something that grows like $\epsilon$ against something that grows like $\epsilon^2$ gives a square root (Figure~\ref{fig:orders}).
Formally, let $\Delta z_t$ be the displacement of the logit vector at position $t$, whatever produced it; stating the argument in logit space covers KV-cache quantization, which does not act on weights.
For small $\Delta z$ the divergence is second order, $D(p_t\|q_t) = \tfrac12\operatorname{Var}_{p_t}(\Delta z_t) + O(\|\Delta z_t\|^3)$, while the displacement of the margin $g_t = \log p_t(i_1) - \log p_t(i_2)$ between the top two tokens, $\Delta g_t = \Delta z_t(i_1) - \Delta z_t(i_2)$, is first order.
Writing
\begin{equation}
\mathbb{E}|\Delta g| = A\sqrt{2\KL}
\label{eq:aniso}
\end{equation}
defines an anisotropy factor $A$ (expectations run over positions with $g_t<1$~nat), and a flip occurs when $\Delta g_t < -g_t$; if displacements are symmetric in sign, independent of the margin and small against the spread of margins,
\begin{equation}
\flips \approx \rhoc\,\mathbb{E}|\Delta g|/2, \qquad \kappa \approx \rhoc\,A/\sqrt2,
\label{eq:firstorder}
\end{equation}
where $\rhoc$ is the density of margins at zero (excluding exact ties): the fraction of positions arbitrarily close to changing their decision, per nat of margin.
The factor $1/2$ is the sign symmetry: only a shift toward the boundary can flip a decision, and compressors push margins down as often as up (the share of negative shifts at fragile positions has median $0.504$ over 372 configurations, with no family departing; App.~\ref{app:derivation}). Dependence of $|\Delta g|$ on $g$ is absorbed in the measured error of the prediction.
The exponent $1/2$ is therefore inherited; what must be measured is where the quadratic regime holds and whether $A$ is comparable across perturbations.
An isotropic noise ladder confirms the quadratic regime below $0.25$ nats (App.~\ref{app:derivation}).
Figure~\ref{fig:regime} shows where it ends: $\flips/\sqrt{\KL}$ is flat up to about $0.25$ nats (from 8 to 3 bits per weight on the rounding ladder), rises above it as higher-order terms enter, and at 2-bit weights, where 94--100\% of decisions flip, saturates, so no power law in KL can continue; flips per unit of total variation stay at $0.75$--$0.98$ of their window value there (App.~\ref{app:beyond}).
In practice: from $10^{-3}$ to $0.25$ nats both statistics read as described here, which covers 8 to 5 bits and most 4-bit models (median $0.108$ nats, but up to $0.75$) on the rounding ladder; up to about 2 nats the square root increasingly underestimates flips ($1.21$ times the window value above 1 nat) while total variation still reads them; beyond that, as the law breaks at 2 bits, only total variation stays within $25\%$.

\paragraph{The exponent.}
Per-reference $R^2\geq0.990$ says little (\S\ref{sec:setup}); the claim rests on the exponents (Table~\ref{tab:exponents}, App.~\ref{app:complementary}) and on the held-out tests.
The exponent is precise (median $95\%$ half-width $\pm0.014$) and, for 30 of 52 references, reliably different from $1/2$; it is ordered by corpus, above $1/2$ on GSM8K and below it on natural text.
That ordering survives removing bf16 ties (Table~\ref{tab:exponents}, last column; App.~\ref{app:derivation}).
The corpus ordering is partly accounted for by the shape of the margin density near zero, in order though not in level (App.~\ref{app:derivation}): the density falls away from zero in every reference, which should push every exponent below $1/2$, yet 26 of 42 exceed it. An exact budget of the exponent resolves this (App.~\ref{app:derivation}): the term of the top-two crossing falls below $1/2$ as the margin shape predicts and sets the corpus order, while flips in which a third token overtakes the leader grow faster than $\sqrt{\KL}$ in every reference and lift the level.

\paragraph{One curve for mechanically unrelated perturbations.}
If divergence were a complete summary of decision damage, the family of a method would add nothing once KL is fixed.
We measure each family's deviation from its reference's power curve (Table~\ref{tab:families} and Figure~\ref{fig:families}, App.~\ref{app:complementary}).
Deviations stay within $5\%$ in sample and within $5.5\%$ when the family is held out of the fit; after a Holm correction over the nine families only layer removal ($0.945$) remains significant, in line with its lower anisotropy factor $A$ (below).
Because the families occupy different parts of the divergence range---KV-cache quantization at low KL, pruning mostly above $0.1$---a family could be judged against a curve from a different regime; restricting the comparison to the window where all families coexist leaves eight families within $1.6\%$ and layer removal at $0.936$ ($95\%$ CI $[0.899, 0.974]$).
The agreement survives stratifying by corpus, dropping \textsc{mix} or the rounding family, changing the calibration corpus, and calibrating on a single family (App.~\ref{app:robust}).

\paragraph{Alignment with the flip directions.}
Equation~\eqref{eq:firstorder} reduces the question to whether $A$ is comparable across families.
It is nearly constant within a reference (median coefficient of variation $2.6\%$), and isotropic Gaussian noise of the same reference supplies a null: every structured compressor's $A$ lies within $0.915$--$1.005$ of the null's, and the two families that differ, layer removal and KV-cache quantization, fall \emph{below} it (App.~\ref{app:robust}).
Standard compressors therefore perturb the decision margins like a generic direction of the same Fisher norm.

\textbf{Answering RQ1.}
Flips grow as a power of KL with exponents near, but not equal to, $1/2$, and within each reference nine families share one curve to within $5.5\%$; they are interchangeable for how many decisions change, not necessarily for what those changes cost (\S\ref{sec:discussion}).

\section{Where the Conversion Factor Comes From}
\label{sec:kappa}

In brief: KL's conversion factor moves with the corpus because KL is averaged over tokens before its square root is taken; total variation has no such average inside it and converts into flips at a ratio near one.

\paragraph{(1) $\kappa$ belongs to the pair (model, corpus).}
Across the 54 references $\kappa$ spans $0.130$--$0.565$; within the 9 models scored on all four generated-text corpora the corpus alone moves it by $1.6$--$2.6\times$, and the ordering GSM8K $<$ \textsc{mix} $<$ MMLU-en $<$ WikiText holds for all 9 models.
Machine translation does not create this effect: $\kappa$ on the Portuguese MMLU half of \textsc{mix} is a median $1.04$ times $\kappa$ on native MMLU-en.

\paragraph{(2) An exact decomposition.}
Because KL is averaged over positions before the square root is taken, $\kappa$ depends on how the divergence is distributed across tokens.
For every configuration,
\begin{equation}
\frac{\flips}{\sqrt{\KL}} = \frac{\flips}{\TV}\cdot \underbrace{\frac{\mathbb{E}_t[\TV_t]}{\mathbb{E}_t[\sqrt{\KL_t}]}}_{c}\cdot \underbrace{\frac{\mathbb{E}_t[\sqrt{\KL_t}]}{\sqrt{\mathbb{E}_t[\KL_t]}}}_{J},
\label{eq:jensen}
\end{equation}
where $\TV_t$ and $\KL_t$ are the per-position terms of Eqs.~\eqref{eq:kl}--\eqref{eq:tv}.
$c$ is a per-token conversion between two first-order quantities, which Pinsker's inequality bounds by $1/\sqrt2$; $J\leq 1$ is a Jensen factor that equals one when every token carries the same divergence and falls as the divergence concentrates on few tokens.
Over the 42 fp32 references (Table~\ref{tab:mechanism}), $c$ is nearly constant (coefficient of variation $4\%$, range $0.44$--$0.51$) while $J$ ranges from $0.31$ to $0.82$.
$\log J$ carries $70\%$ of the variance of $\log r_{0.25}$, against $20\%$ for $\log(\flips/\TV)$ and $11\%$ for $\log c$ (Figure~\ref{fig:kappaj}), a share that survives leaving out any model or corpus (App.~\ref{app:robust}).
The first-order count of Eq.~\eqref{eq:firstorder} agrees with the measured flip rate in aggregate (observed over predicted $0.93$--$1.09$ per corpus), with a small error that is systematic rather than sampling noise (App.~\ref{app:derivation}). The corpus dependence of $\kappa$ is therefore mostly a property of KL as an average, not of the decision geometry: on GSM8K, where most tokens are near-certain and a few carry the divergence, $J$ is $0.38$; on natural text, where uncertainty is spread, it is $0.75$.

\begin{table}[!htb]
  \centering
  \small
  \setlength{\tabcolsep}{3pt}
  \renewcommand{\arraystretch}{0.9}
  \begin{fittable}
  \begin{tabular}{lrrrrrrr}
    \toprule
    corpus & $r_{0.25}$ & $J$ & $\flips/\TV$ & $R_{\text{flat}}$ & $P_{\text{pair}}$ & $A$ & obs/pred \\
    \midrule
    \textsc{gsm8k}        & 0.18 & 0.38 & 1.00 & 1.32 & 0.80 & 3.66 & 1.09 \\
    \textsc{mix}          & 0.22 & 0.45 & 1.06 & 1.64 & 0.70 & 2.72 & 0.95 \\
    \textsc{mmlu-en}      & 0.26 & 0.51 & 1.07 & 1.55 & 0.72 & 2.46 & 1.00 \\
    \textsc{wikitext}     & 0.33 & 0.61 & 1.11 & 1.79 & 0.63 & 2.06 & 0.99 \\
    \textsc{wikitext-nat} & 0.46 & 0.75 & 1.24 & 3.87 & 0.37 & 1.28 & 0.93 \\
    \bottomrule
  \end{tabular}
  \end{fittable}
  \caption{Medians per corpus over the 42 references with an fp32 output head (configurations with $10^{-3}<\KL<0.25$). $r_{0.25}$: median of $r_i$ over this wider window, distinct from $\kappa$ of Eq.~\eqref{eq:kappa} ($\KL<0.05$). $J$: Eq.~\eqref{eq:jensen}. $R_{\text{flat}}$ and $P_{\text{pair}}$: Eq.~\eqref{eq:tvratio}. $A$: Eq.~\eqref{eq:aniso}. obs/pred: measured flip rate over the parameter-free prediction of Eq.~\eqref{eq:firstorder}.}
  \label{tab:mechanism}
\end{table}

\paragraph{(3) Total variation needs no coefficient.}
Total variation is first order at every token, so it avoids the Jensen factor altogether (Figure~\ref{fig:teaser}a).
Refitting every reference against it gives an exponent of $0.971$.
The ratio $\flips/\TV$ has median $1.049$ over all 802 configurations, $95\%$ of them between $0.89$ and $1.29$, and a coefficient of variation of $9.4\%$, against $34\%$ for $\flips/\sqrt{\KL}$.
Averaged per model it is $1.075$ ($95\%$ CI $[1.04, 1.11]$)---close to one, not equal to it---and it rises modestly on high-entropy text, from $1.00$ on GSM8K to $1.21$ on natural text over all configurations (Table~\ref{tab:mechanism} gives the fp32 medians).
Comparing the two statistics fairly means giving each a per-reference constant: total variation then predicts a held-out configuration's flip rate with a median error of $1.6\%$, against $2.4\%$ for $\sqrt{\KL}$, so within one reference the two are nearly interchangeable.
Across references, where a constant cannot be carried over, the split is by order, not by statistic: the per-reference ratio spans $1.33\times$ for total variation and $1.54\times$ for Hellinger distance, both first order at every token, against $3.6\times$ for $\sqrt{\mathrm{JS}}$ and $4.0\times$ for $\sqrt{\KL}$; total variation is the tightest and the only one near one.
Per-token $\sqrt{\KL_t}$, averaged after the root, escapes the Jensen factor by construction: by Eq.~\eqref{eq:jensen}, $\flips/\mathbb{E}_t[\sqrt{\KL_t}] = (\flips/\TV)\cdot c$.
For sampling, total variation is exactly the disagreement probability of the optimal coupling; the arg-max is not a coupling, so the near-unit ratio is not that identity.

\paragraph{Reading two reports.}
Over all pairs of the 802 configurations whose flip rates differ by at least $10\%$, KL assigns the smaller divergence to the configuration that changes more decisions in $0.2\%$ of pairs within one reference, but in $8.1\%$ across corpora of one model and $10.8\%$ across models and corpora ($95\%$ CI $[9.0, 12.4]$, bootstrap over models); total variation does so in $0.0\%$, $0.7\%$ and $1.3\%$.
Within a report KL ranks compressors correctly; across reports it does not.

\paragraph{What total variation guarantees.}
One statement holds without approximation.
\textbf{Lemma 1.} \emph{If the arg-max changes at position $t$, then $\TV_t \geq \delta_t/2$, where $\delta_t = p_t(i_1)-p_t(i_2)$ is the dense model's probability margin. Hence, for every $h>0$,}
\begin{equation}
\flips \;\leq\; F(h) + 2\,\TV/h,
\label{eq:bound}
\end{equation}
\emph{where $F(h)$ is the fraction of positions with $\delta_t<h$.}
The mass $i_1$ loses and the mass the new arg-max gains must together cover the margin, and neither exceeds $\TV_t$ (proof in App.~\ref{app:derivation}).
The bound uses only the dense model and the mean total variation, and it permits far more than compressors produce: if $F(h)\leq\rho h$ near zero it gives $\flips \leq 2\sqrt{2\rho\,\TV}$, a square root of total variation rather than a proportion.
Measured on the 42 fp32 references, the lemma holds at all $2.76$ million flipped positions and no configuration exceeds the bound, which grows as $\TV^{0.52}$ (median over references) while flips grow in proportion; configurations reach a median $16\%$ of it.
The near-unit ratio is therefore not implied by total variation; it describes how compressors spread their divergence, which the rest of this section explains.

\paragraph{Why the ratio is near one.}
Consider a position where only the top two tokens compete, with pair mass $m = p(i_1)+p(i_2)$ and margin $g$.
To first order, $\TV_t = m\,s(g)\,|\Delta g_t|$ with $s(g)=\sigma(g)\sigma(-g)$, and $\int_0^\infty s(g)\,dg = \tfrac12$ exactly.
With a margin density flat near zero and $m=1$, $\mathbb{E}[\TV] = \rhoc\,\mathbb{E}|\Delta g|/2$---the flip count of Eq.~\eqref{eq:firstorder}---so the ratio would be one.
Real distributions depart from that case in two opposite ways: the margin density falls away from zero ($R_{\text{flat}}$ of Eq.~\eqref{eq:tvratio}, $1.04$--$4.48$ over references) and tokens outside the top pair also carry total variation ($P_{\text{pair}}$, $0.32$--$0.84$); the two compensate (Spearman $-0.94$), because a flatter output distribution, as on natural text, spreads both the margins and the moved mass, so their product stays within $0.84$--$1.51$ (App.~\ref{app:derivation}).

A first-order model accounts for this compensation without fitting anything. Isotropic logit noise is an approximation suggested by \S\ref{sec:law}, where compressors align with the flip directions no more than isotropic noise does, and for it the ratio has an exact small-noise limit.
\textbf{Proposition 1.} \emph{Let the logit displacement at every position be i.i.d.\ $\mathcal{N}(0,\sigma^2)$ across a finite set of tokens, and let the gaps between the top token and the others have a bounded joint density whose top-two marginal is continuous at zero with value $\rhoc$. Then}
\begin{equation}
\begin{gathered}
\lim_{\sigma\to0}\frac{\flips}{\TV} = \frac{\sqrt2\,\rhoc}{\mathbb{E}_t[h(p_t)]},\\
h(p) = \textstyle\sum_i p_i\sqrt{1-2p_i+\sum_j p_j^2}.
\end{gathered}
\label{eq:isotropic}
\end{equation}
Here $h(p)/\sqrt{2\pi}$ is the total variation per unit of logit noise at $p$, to first order, so the limit compares margins near zero with the mass noise moves; it is a property of the dense model alone (proof in App.~\ref{app:derivation}); it reduces to $R_{\text{flat}}$ when two tokens compete and tends to one as $T\to0$.
Evaluated on the measured references, the limit overestimates the ratio by a median $9\%$, and the bias comes from linearizing and from extrapolating $\rhoc$, not from isotropy: computing flips and total variation exactly under isotropic Gaussian logit noise matched to each configuration's KL, from the dense distributions alone, predicts the per-reference ratio with a median error of $+3.2\%$, and $-0.5\%$ on the held-out code corpus of \S\ref{sec:holdout} (App.~\ref{app:derivation}).
The near-unit ratio is thus a property of the dense model's output distributions, which compressors inherit to the extent that they displace logits like isotropic noise, as they do in \S\ref{sec:law}. It is also anchored at zero temperature: dividing both models' logits by $T$ leaves every flip unchanged and sends $\flips/\TV\to1$ as $T\to0$; measured, the median stays between $1.01$ and $1.11$ for $T$ from $0.25$ to $2$ (App.~\ref{app:temperature}).

\textbf{Answering RQ2.}
The conversion from KL to flips is set mostly by how concentrated the divergence is across tokens, to which KL, averaged before its square root, is sensitive and total variation is not; the near-unit ratio of flips to total variation is a property of the dense distributions, which isotropic noise reproduces to within $4\%$.
The decomposition is an identity and its variance shares are descriptive; the one experimental manipulation of its factors is the temperature rescaling, which moves $\flips/\TV$.

\section{Out of Sample: New Domains, Models and Kernels}
\label{sec:holdout}

\paragraph{Registered predictions on a code corpus.}
We registered five predictions for three models before generating a sixth corpus (Python solutions to MBPP tasks \citep{austin2021mbpp}, the same twelve configurations), and the same criteria, unchanged, for five more models before generating theirs; each corpus was scored once (App.~\ref{app:holdout}, Table~\ref{tab:holdout}).
The corpus is new; the models are not---all eight are among the 19 of the main study---so this tests transfer to a new domain.
Two predictions held in all eight models: the exponent ($0.48$--$0.57$) and the near-unit ratio of flips to total variation ($0.98$--$1.05$), the only registered prediction about total variation. Its registered tolerance was the published range ($0.76$--$1.29$), which is wide; the observed spread is the informative part.

\paragraph{Where KL fails.}
The prediction of $\kappa$ from $J$ missed its $15\%$ tolerance in five of eight models, because $c$ fell with $J$ on code, and $c$ itself stayed in its registered range in only four.
The registered test that every family stay within $5.5\%$ of the KL curve failed in all eight models; it is stricter than our claim, which concerns model-clustered means, and would have passed only $38\%$ of the published references.
Family deviations were three times those on the published corpora (median $7.5\%$ against $2.4\%$, up to $43\%$), and a comparison we did not register locates the failure in KL: measured against total variation, the same deviations are $2.7\%$ ($90$th percentile $7.0\%$, against $23\%$).
The mechanism is the Jensen factor, which varies across families on code more than three times as much as $\flips/\TV$: fifty-percent sparsity spreads its divergence more evenly across tokens ($J$ $1.28$--$1.53$ times the other families', at a flips-to-TV ratio of $0.96$), so KL understates its flips; rounding, which sits above the curve with a lower $J$, is the exception (App.~\ref{app:holdout}).
Post hoc, then, families stay interchangeable on the new domain in total variation, not in KL.

\paragraph{Mixture of experts and quantization-aware training.}
Applying our criteria post hoc to two MoE models (OLMoE-1B-7B and Qwen3-30B-A3B) and one quantization-aware training family (Gemma-3 Q4\_0) shows the main patterns hold (App.~\ref{app:ext}). In the MoE models, all 8 references meet both criteria (exponents $0.49$--$0.52$, ratios $1.03$--$1.21$); while routing changes carry $7$--$15\%$ of the flips at 4 bits, freezing the router moves the flips-to-TV ratio by only $1.5\%$. QAT checkpoints meet both criteria and fall on the dense model's KL curve; QAT lowers KL to $0.29$--$0.57$ of post-training quantization's, but the flip rate falls only to $0.55$--$0.77$, which the square root predicts to within a median $4\%$: a gain reported in KL overstates the gain in decisions.

\paragraph{A second registered test, in the wild.}
Before generating any data we registered five criteria for three models outside the study (Qwen2.5-14B, Mistral-Nemo-12B and Granite-3.3-8B) on four corpora: the twelve simulated configurations, bitsandbytes NF4 and int8 and FP8 running their real kernels, and Qwen's released AWQ and GPTQ checkpoints (App.~\ref{app:wild}, Table~\ref{tab:wild}).
The exponent stayed in the published range in all 12 references, and the flips-to-TV ratio of real-kernel and released checkpoints stayed in the band of the published configurations in 37 of 38 cases.
The criteria on the level of the ratio failed where it fell below one: on code for Mistral-Nemo (median $0.91$) and, marginally, Granite ($0.987$ against a registered floor of $0.99$), and in the spread of Mistral-Nemo's configurations on code and natural text.
Carried across corpora of the same model, total variation predicted flips with median errors of $9.5$--$15.6\%$ against $34$--$53\%$ for KL, but exceeded the registered $10\%$ for Mistral-Nemo.
FP8 produced non-finite logits in 10 of 12 runs, on which no statistic can be computed; the registration excludes them as kernel failures, so FP8 is untested.

\textbf{Out of sample.}
Total variation keeps its near-unit ratio on a new domain, in two model classes the design left out and in three new models, where it can fall about $10\%$ below one on code; on the new domain, KL keeps its square-root law but not its conversion factor or its shared family curve.

\section{Speculative Decoding: the Exact-Match Consumer}
\label{sec:spec}

\begin{table}[t]
  \centering
  \small
  \setlength{\tabcolsep}{3pt}
  \renewcommand{\arraystretch}{0.9}
  \begin{fittable}
  \begin{tabular}{llrr@{\hskip 10pt}rr}
    \toprule
    & & \multicolumn{2}{c}{self-draft ($n=23$)} & \multicolumn{2}{c}{cross-model ($n=20$)} \\
    stage & predictor & $R^2$ & MAPE & $R^2$ & MAPE \\
    \midrule
    1: flip rate & $\TV$                         & 0.94 & 4.3\% & 0.94 & 3.6\% \\
                 & $\kappa\sqrt{\KL}$, task corpus  & 0.91 & 9.3\% & 0.64 & 12.5\% \\
                 & $\kappa\sqrt{\KL}$, full corpus  & 0.89 & 16.6\% & $-1.96$ & 43.5\% \\
    \midrule
    2: acceptance & i.i.d.\ rounds, measured flips & 0.95 & 2.9\% & 0.79 & 2.9\% \\
                  & i.i.d.\ rounds, $1-\TV$          & 0.95 & 2.5\% & 0.85 & 2.4\% \\
                  & TF agreement sequence           & 0.99 & 1.2\% & 0.98 & 0.7\% \\
    \midrule
    sampled acceptance & $1-\TV$ (TF)                  & 0.95 & 1.3\% & 0.89 & 1.0\% \\
    \bottomrule
  \end{tabular}
  \end{fittable}
  \caption{Predicting speculative decoding (43 target--draft pairs, GSM8K prompts; greedy except the last row). Stage 1 predicts the draft's teacher-forcing flip rate on the task positions from a single scalar; stage 2 converts a flip rate into acceptance per proposed token. The last row predicts the mean acceptance probability of sampled speculative decoding from total variation measured under teacher forcing.}
  \label{tab:spec}
\end{table}

Speculative decoding prices decision fidelity directly (\S\ref{sec:intro}). We ran 23 compressed self-drafts \citep{tiwari2025quantspec,zhao2025qspec} and 20 cross-model drafts (smaller same-family models that share the target's tokenizer, uncompressed and 4-bit, some across model generations) over ten targets on GSM8K prompts, and separate two sources of error (Table~\ref{tab:spec}; Figure~\ref{fig:spec} plots the end-to-end prediction).

In stage 1, total variation predicts the draft's flip rate on the task positions with a mean relative error of $4.3\%$ for self-drafts and $3.6\%$ for cross-model drafts, while $\kappa\sqrt{\KL}$ inherits the corpus dependence of \S\ref{sec:kappa}: calibrated on the target's full reference corpus it errs by $17\%$ and $44\%$, on the task by $9\%$ and $12\%$.
In stage 2, a model of independent rounds converts the \emph{measured} flip rate into acceptance and overestimates it for all 43 pairs, by $2.2$ and $2.3$ points, because disagreements cluster; this bias is common to every scalar, and the full teacher-forcing agreement sequence removes it ($R^2 = 0.99$ and $0.98$).
Under sampling the identity is exact per position, so what is tested is transfer: total variation measured under teacher forcing predicts the acceptance probability at the positions sampled decoding visits ($R^2 = 0.95$ and $0.89$; mean relative errors $1.3\%$ and $1.0\%$).

\paragraph{In a serving stack.}
We then ran speculative decoding in vLLM with real kernels for two Qwen3 targets (14B, 8B) and seven and five drafts (smaller Qwen3 models and their released AWQ and GPTQ versions) on 64 GSM8K and 64 MBPP prompts disjoint from the calibration text, predicting acceptance per proposed token from teacher-forced statistics on that task through the model of independent rounds (Table~\ref{tab:specreal}).
Total variation predicted greedy acceptance with mean errors of $1.1$--$2.4\%$ and ranked the drafts at Spearman $0.86$--$1.00$; $\kappa\sqrt{\KL}$, with $\kappa$ from the target's published references (GSM8K for Qwen3-14B, \textsc{mix} otherwise), erred by $2.5$--$5.1\%$ on GSM8K and $18$--$22\%$ on code, and ranked them at $0.60$--$0.86$. Under sampling the errors were $1.2$--$7.7\%$ for total variation and $3.1$--$22.7\%$ for KL, which did better on both GSM8K targets and far worse on code.
Neither statistic picks the fastest draft reliably: through a cost model proportional to weight bytes, the draft chosen by total variation reached $85$--$100\%$ of the best measured speedup (up to $2.48\times$) and the one chosen by KL $93$--$100\%$, because quantized kernels do not cost in proportion to their bytes.
The round model overestimates acceptance for every pair, so it serves to compare drafts; absolute acceptance and speedup are better read from the teacher-forced agreement sequence.

\textbf{Answering RQ3 (speculative decoding).}
Total variation measured under teacher forcing predicts greedy agreement and sampled acceptance, also in a serving stack, without the corpus-specific calibration that KL requires; choosing the fastest draft further needs a cost model of the kernels.

\section{Discussion and Scope}
\label{sec:discussion}

\paragraph{What to report.}
\begin{itemize}[itemsep=1pt,topsep=2pt,leftmargin=*]
\item \textbf{The consumer needs exact agreement} (speculative drafts, cached or deduplicated outputs, regression tests): report the flip rate itself on the task prompts, and screen drafts by total variation, which predicted acceptance in vLLM with mean relative errors of $1.1$--$2.4\%$ on GSM8K and code, where KL erred by $2.5$--$5.1\%$ and $17.8$--$22.3\%$ (Table~\ref{tab:specreal}); picking the fastest draft also needs a cost model of the kernels.
\item \textbf{Results are compared across corpora or reports}: report total variation, per corpus. KL misorders compressors in $8.1\%$ of such pairs across corpora of one model and $10.8\%$ across models and corpora, total variation in $0.7\%$ and $1.3\%$ (\S\ref{sec:kappa}).
\item \textbf{Methods are compared within one corpus and one dense reference}: KL ranks them correctly ($0.2\%$ of pairs misordered); total variation next to it gives the scale in decisions.
\end{itemize}
When both models can be run on the task, counting flips costs no more than a divergence and should be preferred; total variation matters when a number is read rather than measured, as across reports. Both divergences cost the same teacher-forced pass; a top-$k$ truncation with aggregated tail can only understate total variation (for $k=64$, median $0.1\%$, 95th percentile $7\%$), and bf16 logits need an fp32 output head against ties. The released code reports all three for any dense--compressed pair.

\paragraph{Decisions are not accuracy.}
None of these statistics measures what a changed decision costs: on Gemma-3-4B at a matched GSM8K divergence of $0.10$ nats and nearly the same flip rate, magnitude pruning costs $33.3$ accuracy points while Wanda gains $4.8$. A flip rate or a divergence describes fidelity to the dense model, not benchmark quality. Nor do per-position statistics say how changed decisions compound over free generation.

\paragraph{Scope.}
The main study uses dense decoder-only transformers and post-training changes of fixed weights or of the KV cache. Mixture-of-experts models and quantization-aware training enter only as a post hoc extension with two MoE models and one QAT family (App.~\ref{app:ext}); larger MoE models, quantized routers, other QAT recipes, fine-tuning in general, distillation and multimodal inputs are untested. Every statistic is measured under teacher forcing on greedy references. What should carry over is what Proposition~1 needs: perturbations that move logits like isotropic noise; where they do not, as partly for layer removal, departures should be expected, so adapters or fine-tuning must be checked, not assumed. All statements hold inside the measured window ($10^{-3}$ to about 2 nats; 8 to 3 bits per weight; 1B--72B parameters; greedy decoding; five corpora in two languages and one held-out code corpus), within the ranges of \S\ref{sec:law}.

\section*{Limitations}
\label{sec:limitations}
FP8 was not tested (its real kernel produced non-finite logits in most runs). The number of independent units is modest (54 references, several corpora sharing models); the speculative analyses rest on ten targets with 43 drafts, and the serving test on two targets. $p$-values other than the Holm-corrected family tests are descriptive.
Two statements are proved, the bound of Lemma~1 and the isotropic small-noise limit of Proposition~1; real compressors are neither isotropic nor infinitesimal, so the near-unit ratio is explained, not guaranteed, and the budget of the exponent locates its departure from $1/2$ in third-token crossings without modelling their rate (App.~\ref{app:derivation}).\label{pg:endcontent}

\makeatletter\@ifpackageloaded{acl}{}{\bibliographystyle{acl_natbib}}\makeatother
\bibliography{paper}

\appendix

\section{Replication Details}
\label{app:replication}

\subsection{Evaluation hygiene}
\label{app:hygiene}
\paragraph{bfloat16 ties.}
In bf16 the extrapolation that defines $\rhoc$ fails structurally: between the atom of exact ties and the first grid step there are no margins, so the density is zero out to $h=0.1$ and the estimate is negative for 50 of the 54 bf16 references (Figure~\ref{fig:density}); every margin analysis therefore uses an fp32 output head.
\paragraph{Degenerate reference corpora.}
Greedy continuation of raw text makes instruction-tuned models loop: in a diagnostic run, 99--100\% of Gemma sequences and 51\% of Qwen3-4B sequences repeated more than half of their 4-grams, $\kappa$ fell by an order of magnitude and the exponent rose to $0.85$ ($R^2 = 0.56$); those corpora were discarded. With the excerpt inside the chat template the exponent returned to $0.465$, and the pipeline now refuses corpora with more than $10\%$ degenerate sequences; stored generated corpora are $0$--$0.4\%$ degenerate for instruction-tuned models and $31\%$ for the one base model generated greedily, which is why \textsc{wikitext-nat} exists.
\subsection{Derivations}
\label{app:derivation}
\paragraph{Proof of Lemma 1.}
Let the arg-max of $q_t$ be $j\neq i_1$, so $q_t(j)\geq q_t(i_1)$. Then $\delta_t \leq p_t(i_1)-p_t(j) \leq [p_t(i_1)-q_t(i_1)] + [q_t(j)-p_t(j)]$. The mass lost by the tokens whose probability falls sums to $\TV_t$, and so does the mass gained by the tokens whose probability rises; each bracket is at most one of these sums, so $\delta_t\leq 2\TV_t$. For Eq.~\eqref{eq:bound}, split positions at $\delta_t<h$: those with $\delta_t\geq h$ can flip only if $\TV_t\geq h/2$, and by Markov's inequality their share is at most $2\,\TV/h$. The measured check uses total variation on the dense top-64 plus an aggregated tail, which can only understate the full value; the lemma still held at every flipped position.

\paragraph{Second order in weights.}
For a weight perturbation $\delta$, the divergence is minimized at $\delta=0$, so $\KL = \tfrac12\delta^\top F\delta + O(\|\delta\|^3)$ with $F$ the Fisher information \citep{amari1998natural,martens2020fisher}, while $\Delta g_t = \nabla_\theta g_t\cdot\delta + O(\|\delta\|^2)$; the logit-space form of \S\ref{sec:law} is the case $\Delta z = J_\theta\delta$ and also covers perturbations of activations. On the isotropic noise ladder, measured KL over $\tfrac12\sigma^2\operatorname{tr}F$ has median $0.986$ inside the $\kappa$ window (at most $11\%$ off, 126 configurations over 15 references; at most $15\%$ off below $0.25$ nats) and $1.40$ beyond 2 nats.
\paragraph{Sign symmetry.}
Eq.~\eqref{eq:firstorder} assumes $\Delta g$ symmetric in sign. Over fragile positions of 372 configurations ($10^{-3}<\KL<0.25$), the proportion with $\Delta g<0$ has median $0.504$ (5th--95th percentile $0.488$--$0.523$), with no family departing. Recomputing the prediction with the fragile threshold at $0.5$, $1$ and $2$ nats moves the median observed-to-predicted ratio by at most $1.4\%$ (34 references).
\paragraph{Margin shape and the exponent.}
The corpus ordering of the exponent is not produced by bf16 ties, which could flatten the curve at low divergence: it is unchanged on bf16 references restricted to $\KL>10^{-2}$, and on the fp32 references it keeps natural text lowest and GSM8K highest, with \textsc{mix} and MMLU-en tied. With $\rho(g)\approx\rho_0+\rho_1 g$ near the boundary, the flip rate gains a term in $\rho_1\KL$, so references whose margin density falls less steeply from zero should show larger exponents. They do: fitting $\rho_1/\rho_0$ on the margins below 1~nat of the 42 fp32 references, the exponent orders with it at Spearman $0.86$ ($0.62$ within corpora). The prediction holds in order but not in level---the density falls from zero in every reference, which should push every exponent below $1/2$, yet 26 of 42 exceed it.
An exact budget locates the level. With $\rhoc$ fixed per reference, $\flips = \eta\,\rhoc A\sqrt{2\KL}/2$, where $\eta$ is the observed-to-predicted ratio, and $\eta = \eta_{\text{pair}}/\text{cover}$, where cover is the share of flips that are top-1/top-2 crossings; regressing the logs on $\log\KL$ over the same configurations therefore splits the fitted exponent exactly into $\alpha = \tfrac12 + \beta_A + \beta_{\text{pair}} + \beta_{\text{out}}$. Over the 42 fp32 references ($10^{-3}<\KL<2$), $A$ barely drifts (median $\beta_A = -0.013$); the pair term is negative in 38, as the margin shape predicts (median $-0.050$), and orders the corpora from $-0.008$ on GSM8K to $-0.096$ on natural text; the term of flips in which a third token overtakes the leader is positive in all 42 (median $+0.065$). The two compensate, which is why exponents straddle $1/2$ although the density falls from zero in every reference. The budget is exact but descriptive: it locates the excess in third-token crossings without modelling their rate.

\paragraph{Beyond the quadratic regime.}
\label{app:beyond}
The Fisher prediction of KL degrades to $60\%$ at the top of the isotropic ladder. Above $0.25$ nats, $A$ stays at a median $0.96$ of its low-divergence value in the same reference, but the local exponent rises from $0.497$ to $0.550$ (21 references). A single quadratic term over the full range is small (mean $+0.004$, $95\%$ CI $[-0.001, +0.009]$, with no consistent sign), and the interval is compatible with the steepening above $0.25$ nats, which is confined to the top of the range; the full-range exponents average over both regimes. On the rounding ladder the window runs from 8 bits per weight (median $\KL = 0.0009$) to 3 bits ($0.340$). Relative to the $\kappa$ window, $\flips/\sqrt{\KL}$ is $1.21$ above 1 nat; in 32 configurations beyond the measured range (2-bit weights, 50--70\% pruning, 2-bit KV cache; four models, entering no fit) it peaks at $1.80$, and at 2-bit weights KL reaches 8--18 nats.

\paragraph{First-order decomposition of $\kappa$.}
Equation~\eqref{eq:firstorder} writes $\kappa \approx \rhoc A/\sqrt2$.
Substituting the definition of $A$ makes this an identity whenever the first-order counting holds, so its content is empirical: the measured flip rate over the parameter-free prediction has median $0.93$--$1.09$ per corpus (Table~\ref{tab:mechanism}). That range is not sampling noise alone: bootstrap intervals over references contain one on three corpora, but \textsc{mix} and natural text sit $5\%$ and $7\%$ below it and the corpora differ (Kruskal--Wallis $p=0.03$), a small systematic error of the first-order count. $A$, which needs one scoring pass over a compressed copy, is nearly constant within a reference while across corpora it moves by $2.9\times$ in the direction opposite to $\rhoc$ (Table~\ref{tab:mechanism}).
The agreement is aggregate rather than per event: the top-2 crossing that Eq.~\eqref{eq:firstorder} counts accounts for a median $87\%$ of flips at $\KL<0.25$ ($92\%$ at $\KL<0.05$).
A cheaper, dense-only proxy---the fraction of tokens with margin below one nat---tracks $\kappa$ closely on generated text ($r = 0.991$ over 45 references) but is not a mechanism: natural-text references sit $6.6$ residual standard deviations below its line, an offset the decomposition removes (below).

\paragraph{Proof of Proposition 1.}
\emph{Flips.} The top-two gap moves by $\Delta g = \Delta z_{i_1}-\Delta z_{i_2}\sim\mathcal{N}(0,2\sigma^2)$, independent of the gap $g$, and the crossing $\Delta g<-g$ has probability $\int\rho(g)\,\Phi(-g/\sigma\sqrt2)\,dg = \sigma\sqrt2\int_0^\infty\rho(\sigma\sqrt2\,u)\,\Phi(-u)\,du$, which tends to $\rhoc\,\sigma\sqrt2\int_0^\infty\Phi(-u)\,du = \rhoc\,\sigma/\sqrt\pi$ by dominated convergence ($\rho$ bounded, $\int_0^\infty\Phi(-u)\,du = 1/\sqrt{2\pi}$). A crossing by any other token $j$ needs its gap $g_j\geq g$ to be of order $\sigma$, so both gaps are; with a bounded joint density this has probability $O(\sigma^2)$ up to Gaussian tails, so $\flips = \rhoc\,\sigma/\sqrt\pi + o(\sigma)$, which is Eq.~\eqref{eq:firstorder} with $\mathbb{E}|\Delta g| = 2\sigma/\sqrt\pi$.
\emph{Total variation.} On a finite set of tokens $q_i - p_i = p_i(\Delta z_i - \sum_j p_j\Delta z_j) + O(\sigma^2)$ uniformly, and the linear term is Gaussian with variance $p_i^2\sigma^2(1-2p_i+\sum_j p_j^2)$, so $\mathbb{E}\,\TV_t = \sigma\,h(p_t)/\sqrt{2\pi} + O(\sigma^2)$; dividing, $\sigma$ cancels and the limit is Eq.~\eqref{eq:isotropic}. In the measured setting, tokens outside the dense top-64 enter $h$ as their total mass times $(1+\sum_j p_j^2)^{1/2}$. For two competing tokens with margin $g$, $h = 2\sqrt2\,\sigma(g)\sigma(-g)$ and the ratio is $R_{\text{flat}}$. Dividing the logits by $T$ scales both $\rhoc$ and $\mathbb{E}[h]$ by $T$ as $T\to0$, so the ratio tends to one. Over the 42 fp32 references the prediction ranges over $0.89$--$1.56$ against a measured $0.98$--$1.30$ (while $R_{\text{flat}}$ ranges over $1.04$--$4.48$), with median error $+8.8\%$, Spearman $0.66$ and maximum error $27\%$; per corpus it gives $1.02$ against a measured $1.00$ on GSM8K and $1.43$ against $1.24$ on natural text.

\paragraph{Exact isotropic model.}
On the dense top-64 distribution plus its aggregated tail, i.i.d.\ Gaussian noise of standard deviation $\sigma$ is added to the 64 logits (the tail is held fixed), and KL, total variation and flips are computed exactly, averaged over every position of the reference, on a grid of 28 values of $\sigma$ from $0.02$ to $3$. Each configuration in $10^{-3}<\KL<0.25$ is read off the grid at its measured KL (log-linear interpolation), so no flip or total-variation measurement enters the prediction; 372 configurations of the 42 fp32 references and 76 of the held-out code corpus, whose references entered no step of the design. The prediction has mean absolute error $4.2\%$ (Pearson $0.95$ in logs) on the published references and $3.0\%$ (at most $5.7\%$) on the held-out code corpus; the residual concentrates on natural text (up to $17\%$; at most $9.6\%$ elsewhere), and the three references with errors above $10\%$ are natural text. At equal KL, isotropic noise yields a median $11\%$ more total variation and $14\%$ more flips than the measured compressors; the two excesses cancel in the ratio.

\paragraph{Dense-only proxy.}
\label{app:proxy}
Replacing $\rhoc$ by the fraction of fragile tokens, $\phi = P(g<1\,\text{nat})$, a line fits the 45 generated-text references well ($\kappa = 1.10\,\phi + 0.114$, $r = 0.991$), and thresholds of $0.5$ and $2$ nats give $r = 0.986$ and $0.992$. The threshold is not critical because it barely changes which references are fragile: the fractions below $0.5$, $1$ and $2$ nats rank the 54 references almost identically (Spearman $0.998$), which changes only the slope of the fit. The intercept has no physical reading (a model with no fragile tokens would flip nothing), and the natural-text references sit $6.6$ residual standard deviations below the line ($6.5$ and $5.5$ at the other thresholds; Figure~\ref{fig:proxy-panel}).

\paragraph{Total variation of a two-way competition.}
The measured ratio factorizes into the two departures from a flat, two-token case,
\begin{equation}
\frac{\flips}{\TV} = \underbrace{\frac{\rhoc}{2\,\mathbb{E}[m\,s(g)]}}_{R_{\text{flat}}}\cdot\underbrace{\frac{\mathbb{E}[m\,s(g)|\Delta g|]}{\mathbb{E}[\TV_t]}}_{P_{\text{pair}}}\cdot\eta,
\label{eq:tvratio}
\end{equation}
where $\eta$ collects the error of Eq.~\eqref{eq:firstorder} and the dependence of $|\Delta g|$ on $g$ (median $0.95$ over references, 10th--90th percentile $0.89$--$1.07$). $R_{\text{flat}}$ exceeds one because the margin density falls away from zero; $P_{\text{pair}}$, the share of total variation carried by the top pair, falls below one because other tokens also move; a flatter output distribution spreads both margins and moved mass.

If only $i_1,i_2$ move, $\TV_t = |\Delta p(i_1)|$, and with $p(i_1)=m\,\sigma(g)$ at fixed pair mass $m$, $\Delta p(i_1)\approx m\,\sigma(g)\sigma(-g)\,\Delta g$. Averaging over positions with $\Delta g$ independent of $g$ gives $\mathbb{E}[\TV] = \mathbb{E}[m\,s(g)]\,\mathbb{E}|\Delta g|$, and with a flat density $\rho_0$ on the scale of $s$, $\mathbb{E}[m\,s(g)] = m\rho_0\int_0^\infty s = m\rho_0/2$. Eq.~\eqref{eq:tvratio} is the exact factorization of the measured ratio; $\eta$ is the product of the measured-to-predicted ratio of Eq.~\eqref{eq:firstorder} and $\mathbb{E}_{g<1}|\Delta g| / \mathbb{E}_{m s}|\Delta g|$.

\subsection{Temperature}
\label{app:temperature}
Dividing the logits of both models by $T$ changes no arg-max, so the flips are fixed and only total variation moves. As $T\to0$ both distributions become one-hot and $\TV_t$ becomes exactly the flip indicator, so $\flips/\TV\to1$ for any perturbation; $R_{\text{flat}}$ and $P_{\text{pair}}$ also tend to one, since the margin density is flat on a vanishing scale and only the top pair carries mass. Measured on the top-64 tokens of the 42 fp32 references (Figure~\ref{fig:temperature}), the median ratio is $1.01$ at $T=0.25$ (range $0.99$--$1.07$) and $1.07$ at $T=1$, and it drifts upward with temperature on high-entropy corpora. At $T=2$, $R_{\text{flat}}$ reaches $13.5$ and $P_{\text{pair}}$ falls to $0.14$, yet the ratio stays within $0.82$--$1.63$ (median $1.11$): the same property of the output distribution---how many tokens compete near the top---inflates one correction and deflates the other. The first-order formula of Eq.~\eqref{eq:isotropic} follows the rise with $T$ but overestimates it, by $+5\%$ at $T=0.25$ to $+14\%$ at $T=2$.

\subsection{Out-of-sample validation}
\label{app:holdout}
The criteria were fixed before the code corpus was generated, from published numbers only, time-stamped and released with the code (\texttt{results/\allowbreak holdout\_\allowbreak predictions.json}, three models). The extension to five more models (\texttt{results/\allowbreak holdout\_\allowbreak predictions\_\allowbreak ext.json}) repeats the criteria unchanged, records the hash of the original file, and was registered before any of their completions were generated. Corpus: 256 MBPP test tasks, completions generated greedily by each model inside its chat template; configurations u8, u6, u5, u4, kv4, kv3, mag20, g4, gptq4, awq4, sgpt50 and wanda50, scored with an fp32 output head; nothing from this corpus enters any other analysis. The five criteria: P1, per-reference exponent within the published range $0.459$--$0.578$; P2, every family within $5.5\%$ of the curve fitted without it; P3, median flips/TV within the published per-reference range $0.76$--$1.29$; P4, per-token conversion $c$ within $0.44$--$0.51$; P5, $\kappa$ predicted from $J$ with the published medians of flips/TV ($1.07$) and $c$ ($0.47$) within $15\%$. Outcomes per model are in Table~\ref{tab:holdout}. P2 as registered is stricter than the claim of \S\ref{sec:law}, which concerns the model-clustered mean: applied to the published fp32 references it would pass only $38\%$ of them; the post hoc comparison in \S\ref{sec:holdout} uses the same per-family, per-reference deviations on both sides.

\paragraph{Post hoc analyses.} The post hoc analyses of this corpus refit each family's deviation with total variation in place of $\sqrt{\KL}$: on code the median falls from $7.5\%$ to $2.7\%$ and the maximum from $43\%$ to $18\%$, half of the eight models then keep every family within $5.5\%$ (none does against KL), and on the 42 published fp32 references the median falls from $2.4\%$ to $1.6\%$. Writing each family's deviation against KL as the sum of its log ratios of $J$, $\flips/\TV$ and $c$ to the other families of the reference explains $67\%$ of its variance over the 64 (model, family) pairs; $J$ varies across families more than three times as much as $\flips/\TV$ (standard deviations $0.18$ and $0.05$ in logs). Rounding is the exception: it sits $11\%$ above the curve with a lower $J$ ($0.88$ times the others'). Recalibrating GPTQ, AWQ, SparseGPT and Wanda on Python files (codeparrot, disjoint from MBPP; 128 windows of 512 tokens) lowers their KL by a median $28\%$, moves their median deviation against KL from $7.7\%$ to $7.3\%$ (maximum $43\%$ to $36\%$), and leaves it at $2.3\%$ against total variation. Also post hoc: median family deviations over the eight models are $+21\%$ (SparseGPT), $+12\%$ (Wanda) and $+11\%$ (rounding), above the curve in 7, 6 and 7 models, against $-9\%$ to $-2\%$ for the other families; across the 64 (model, family) pairs the deviation is unrelated to the family's mean log divergence (Spearman $0.16$, $p=0.21$), and refitting inside $\KL<0.25$ lowers the median absolute deviation to $4.4\%$ but leaves a maximum of $31\%$.

\subsection{The registered test in the wild}
\label{app:wild}
The criteria were written, hashed and saved (\texttt{results/\allowbreak inthewild\_\allowbreak predictions.json}) before any reference corpus or score of the three models was generated. Models: Qwen2.5-14B-Instruct, Mistral-Nemo-Instruct-2407 and Granite-3.3-8B-Instruct, none among the 19 of the main study. Corpora: \textsc{gsm8k}, \textsc{mmlu-en}, \textsc{wikitext-nat} and code. Configurations: the twelve of App.~\ref{app:holdout}; bitsandbytes NF4 and int8 and FP8 loaded through their real kernels; and, for Qwen2.5-14B, the released AWQ, GPTQ-Int4 and GPTQ-Int8 checkpoints, whose published weights are dequantized. W1: per-reference exponent within $0.459$--$0.578$. W2: per-reference median flips/TV within $0.99$--$1.25$, the 5th--95th percentile band of the 54 published per-reference medians. W3: at least $90\%$ of configurations with flips/TV within $0.89$--$1.29$, the central $95\%$ of the 802 published configurations. W4: every real-kernel and released checkpoint within $0.89$--$1.29$. W5: predicting a corpus's flips from another corpus of the same model, total variation errs less than $\sqrt{\KL}$ and at most $10\%$ (median). A run with non-finite KL or total variation counts as a kernel failure and is excluded.

\subsection{Mixture-of-experts and quantization-aware training}
\label{app:ext}
This extension was run after the study, in a separate results tree that no analysis of the main text reads, and was not registered; it applies the registered criteria P1 (exponent within $0.459$--$0.578$) and P3 (median flips/TV within $0.76$--$1.29$) unchanged. Every reference uses an fp32 output head and the twelve configurations of App.~\ref{app:holdout}.
\paragraph{Mixture of experts.} OLMoE-1B-7B-0125-Instruct (64 experts, 8 active) and Qwen3-30B-A3B-Instruct-2507 (128 experts, 8 active) on \textsc{gsm8k}, \textsc{mmlu-en}, \textsc{wikitext} and \textsc{wikitext-nat}, and OLMoE also on the code corpus. The router is left in bf16 and outside every compressor, as in deployed quantizers; in the calibrated methods, an expert that receives no calibration token gets the uncalibrated operator of the same format (rounding or magnitude pruning). P1 and P3 hold in all 8 references on the published corpora (exponents $0.49$--$0.52$, ratios $1.03$--$1.21$) and on code (exponent $0.51$, ratio $0.99$); per reference, the median family deviation is $0.9$--$2.2\%$ against the KL curve and $0.8$--$2.5\%$ against total variation, and on code $10.0\%$ and $3.3\%$.
\paragraph{Routing.} On 64 sequences per reference and nine configurations (81 cells), each configuration is scored twice: as is, and with every router's logits replaced by the reference's, which fixes both the selected experts and their gating weights. The difference in flips is the share carried by routing: $7$--$15\%$ at 4 bits (rounding, Gaussian noise, GPTQ) and $48$--$66\%$ at 8 bits, where $40$--$57\%$ of (layer, position) pairs already change their expert set at 4 bits. Router decisions have thin margins: the median gap between the $k$-th and $(k{+}1)$-th router logit is $0.047$--$0.078$ nats and $62$--$79\%$ of decisions lie within $0.1$ nats. Because the router runs in bf16, whose grid these margins reach, part of the routing share at 8 bits may be exact ties, as for tokens (App.~\ref{app:hygiene}). Freezing routing moves flips/TV by a median $1.5\%$ and by less than $10\%$ in 73 of the 81 cells; with routing, flips/TV spans $0.85$--$1.30$.
\paragraph{Quantization-aware training.} The Gemma-3 checkpoints of 1B, 4B and 12B trained for the Q4\_0 format of llama.cpp (released unquantized) are scored with their own references on the four corpora, and against the dense Gemma-3 references on the same text, which here are rescored with the same configurations plus Q4\_0 (symmetric 4-bit, groups of 32). P1 and P3 hold in all 12 references of each (QAT exponents $0.47$--$0.56$, ratios $0.99$--$1.24$). Against the dense reference, the unquantized QAT checkpoint and its Q4\_0 version deviate from the dense KL curve by factors $1.01$--$1.11$ and from its flips/TV by $0.93$--$1.04$. Relative to Q4\_0 applied to the dense model, QAT's Q4\_0 has $0.29$--$0.57$ of the KL and $0.55$--$0.77$ of the flips; the square root of the KL ratio predicts the flip ratio with a median absolute error of $3.8\%$ (at most $7.0\%$), and the ratio of total variations with $3.8\%$ (at most $7.8\%$).

\subsection{Coverage}
\label{app:coverage}
A 4-bit model exported to a production kernel (AWQ-Marlin in vLLM) changes the simulated top-1 at $0.20\%$ of 1024 positions, as often as two kernels running the dense model. On 4096 positions per configuration scored over the full vocabulary ($n=1312$ configurations), the full KL and TV exceed the truncated ones by a median factor of $1.003$ and $1.001$ (p95 $1.11$ and $1.07$), with no trend in the divergence.
Calibrating on the evaluation corpus does not produce the agreement: recompressing the four calibrated methods with C4 instead of WikiText-2 on 8 WikiText references raises their KL by a median factor of $1.08$ but moves them along the curve rather than off it (deviation factor $1.00$, range $0.94$--$1.03$). One configuration of a family also suffices to calibrate the others: $\kappa$ taken from a single family predicts every other family of the reference with a median error of $2.5\%$ (isotropic noise) to $3.7\%$ (any compressor), against $2.4\%$ when calibrated on all other families; only layer removal calibrates worse ($5.3\%$).

Of the 54 references, 52 have at least 8 configurations and enter the exponent and family analyses (Qwen2.5-72B on \textsc{wikitext-nat} and Qwen3-8B on \textsc{mix} have fewer); 42 were rescored with an fp32 output head and enter every margin analysis (Table~\ref{tab:mechanism}, Eqs.~\eqref{eq:firstorder}--\eqref{eq:tvratio}); the six models without fp32 rescoring are Gemma-3-27B, Qwen2.5-7B and 72B, Mistral-Small-24B, phi-4 and OLMo-2-13B. The Gaussian-noise null of \S\ref{sec:law} uses the 31 fp32 references that carry the Gaussian family; the threshold sensitivity was run on 34 of them; the split at $0.25$ nats uses the 37 references with at least three configurations below and one above for $A$, and the 21 with at least four on each side for the local exponent. The per-family configuration counts are RTN 225, KV-cache 110, Gaussian noise 73, magnitude 66, GPTQ 64, AWQ 63, Wanda 61, SparseGPT 60, single-layer 52, layer removal 24 and third-party 4.

\subsection{Robustness of the family comparison and of the decomposition}
\label{app:robust}
\paragraph{Families.} Stratified by corpus, no family departs from its reference's curve by more than $3.4\%$ outside \textsc{mix}; dropping \textsc{mix} (45\% of the configurations) leaves eight families within $2.9\%$, and fitting each curve without RTN and its single-layer variant leaves every family within $5.1\%$. Neither the calibration corpus nor the choice of calibrating family matters (App.~\ref{app:coverage}).
\paragraph{Gaussian null.} Against the Gaussian null of \S\ref{sec:law}, SparseGPT ($0.945$) and Wanda ($0.996$) rest on 7 references with wide intervals and layer removal ($0.915$) on 3.
\paragraph{Decomposition.} The share of the variance of $\log r_{0.25}$ carried by $\log J$ stays at $69$--$71\%$ leaving out one of the 13 models at a time, $69$--$72\%$ leaving out one corpus, and $69\%$ within models.

\subsection{Reproducibility}
Every number in the paper is produced by the analysis stages of the accompanying code and collected into one file by the figure script; a checker verifies 1181 claims of the manuscript against that file and fails on any mismatch. The stages cover reference construction and scoring, the regression and first-order analyses, the decomposition and the bound of \S\ref{sec:kappa}, the calibration control, speculative prediction, the held-out corpus and its post hoc analyses, the isotropic model, the budget of the exponent, the comparison of rankings across reports, the mixture-of-experts and quantization-aware extension, and the registered test in the wild with its real kernels. The only value not regenerated by the pipeline is the diagnostic degeneration rate of App.~\ref{app:hygiene}, whose corpora were discarded.

\section{Complementary Results}
\label{app:complementary}
\setcounter{figure}{0}\renewcommand{\thefigure}{B\arabic{figure}}
\setcounter{table}{0}\renewcommand{\thetable}{B\arabic{table}}
\renewcommand{\theHfigure}{B\arabic{figure}}\renewcommand{\theHtable}{B\arabic{table}}
\setcounter{topnumber}{4}\setcounter{bottomnumber}{4}\setcounter{totalnumber}{8}
\renewcommand{\topfraction}{0.95}\renewcommand{\bottomfraction}{0.95}\renewcommand{\textfraction}{0.02}
\renewcommand{\floatpagefraction}{0.9}

\begin{table}[!htb]
  \centering
  \small
  \setlength{\tabcolsep}{3pt}
  \renewcommand{\arraystretch}{0.9}
  \begin{fittable}
  \begin{tabular}{lrrrrr}
    \toprule
    corpus & refs & configs & exponent range & mean [95\% CI] & fp32 mean \\
    \midrule
    \textsc{mix}          & 17 & 363 & 0.478--0.563 & 0.514 [0.502, 0.526] & 0.504 \\
    \textsc{gsm8k}        &  9 & 108 & 0.507--0.578 & 0.529 [0.512, 0.547] & 0.524 \\
    \textsc{mmlu-en}      &  9 & 108 & 0.482--0.532 & 0.509 [0.496, 0.522] & 0.504 \\
    \textsc{wikitext}     & 10 & 120 & 0.473--0.516 & 0.496 [0.486, 0.505] & 0.495 \\
    \textsc{wikitext-nat} &  9 & 103 & 0.459--0.495 & 0.478 [0.467, 0.490] & 0.476 \\
    \bottomrule
  \end{tabular}
  \end{fittable}
  \caption{Exponents of the per-reference fit $\log \flips = \alpha \log \KL + \beta$ on bf16 references; the range is over references, the mean and its interval are clustered across references (52 references with at least 8 configurations). The last column refits the 42 references rescored with an fp32 output head, which removes bf16 ties.}
  \label{tab:exponents}
\end{table}

\begin{table}[!htb]
  \centering
  \small
  \setlength{\tabcolsep}{2.5pt}
  \renewcommand{\arraystretch}{0.9}
  \begin{fittable}
  \begin{tabular}{llcccccc}
    \toprule
    & & & \multicolumn{2}{c}{accept.\ error} & \multicolumn{2}{c}{Spearman} & \\
    target & task & $\kappa$ from & TV & KL & TV & KL & best speedup reached (TV / KL) \\
    \midrule
    Qwen3-14B & GSM8K & GSM8K & $1.1\%$ & $2.5\%$  & $1.00$ & $0.86$ & $85\%$ / $100\%$ \\
    Qwen3-8B  & GSM8K & \textsc{mix} & $1.8\%$ & $5.1\%$  & $0.90$ & $0.70$ & $95\%$ / $95\%$ \\
    Qwen3-14B & code  & \textsc{mix} & $2.1\%$ & $17.8\%$ & $0.86$ & $0.61$ & $100\%$ / $100\%$ \\
    Qwen3-8B  & code  & \textsc{mix} & $2.4\%$ & $22.3\%$ & $0.90$ & $0.60$ & $85\%$ / $93\%$ \\
    \bottomrule
  \end{tabular}
  \end{fittable}
  \caption{Speculative decoding in vLLM with real kernels (greedy; 7 drafts for Qwen3-14B, 5 for Qwen3-8B; 64 prompts per task, disjoint from the calibration text). Acceptance error: mean relative error of the predicted acceptance per proposed token. Last column: measured speedup of the draft each statistic selects through a weight-byte cost model, relative to the fastest draft.}
  \label{tab:specreal}
\end{table}

\begin{table}[!htb]
  \centering
  \small
  \setlength{\tabcolsep}{3pt}
  \renewcommand{\arraystretch}{0.9}
  \begin{fittable}
  \begin{tabular}{lccccc}
    \toprule
    model & W1 & W2 & W3 & W4 & W5 (TV / KL error) \\
    \midrule
    Qwen2.5-14B     & 4/4 & 4/4 & 4/4 & 21/21 & pass ($9.5\%$ / $53.2\%$) \\
    Mistral-Nemo    & 4/4 & 3/4 & 2/4 & 8/9   & fail ($15.6\%$ / $51.8\%$) \\
    Granite-3.3-8B  & 4/4 & 3/4 & 4/4 & 8/8   & pass ($9.8\%$ / $34.1\%$) \\
    \bottomrule
  \end{tabular}
  \end{fittable}
  \caption{The registered test in the wild (criteria in App.~\ref{app:wild}). Failures: W2, code for Mistral-Nemo ($0.905$) and Granite ($0.987$); W3, Mistral-Nemo on natural text ($75\%$ inside) and code ($58\%$); W4, Mistral-Nemo's int8 kernel on code ($0.851$). FP8 runs with non-finite logits (10 of 12) are excluded as registered.}
  \label{tab:wild}
\end{table}

\begin{table}[!htb]
  \centering
  \small
  \setlength{\tabcolsep}{3pt}
  \renewcommand{\arraystretch}{0.9}
  \begin{fittable}
  \begin{tabular}{lrrrrr}
    \toprule
    model & P1 exponent & P2 max dev. & P3 flips/TV & P4 $c$ & P5 error \\
    \midrule
    Mistral-7B   & 0.497 & 6.7\%  & 1.01 & 0.46 & $+8\%$ \\
    Qwen3-4B     & 0.519 & 38.5\% & 1.02 & 0.41 & $+16\%$ \\
    Gemma-3-4B   & 0.557 & 21.9\% & 0.98 & 0.43 & $+23\%$ \\
    \midrule
    Llama-3-8B   & 0.565 & 10.5\% & 0.98 & 0.44 & $+19\%$ \\
    OLMo-2-7B    & 0.479 & 17.6\% & 1.05 & 0.46 & $+5\%$ \\
    Phi-3.5-mini & 0.531 & 32.6\% & 0.99 & 0.44 & $+14\%$ \\
    Qwen3-8B     & 0.574 & 20.8\% & 1.00 & 0.41 & $+25\%$ \\
    Gemma-3-12B  & 0.559 & 42.6\% & 0.99 & 0.43 & $+20\%$ \\
    \midrule
    passed       & 8 of 8 & 0 of 8 & 8 of 8 & 4 of 8 & 3 of 8 \\
    \bottomrule
  \end{tabular}
  \end{fittable}
  \caption{Registered predictions on the held-out code corpus (criteria in App.~\ref{app:holdout}). The first three models were registered before the corpus was generated, the other five before their completions were generated; all eight are also in the main study.}
  \label{tab:holdout}
\end{table}

\begin{table}[!htb]
  \centering
  \small
  \setlength{\tabcolsep}{3pt}
  \renewcommand{\arraystretch}{0.9}
  \begin{fittable}
  \begin{tabular}{lrrrr}
    \toprule
    family & in-sample & held out & 95\% CI (held out) & overlap \\
    \midrule
    layer removal    & 0.950 & 0.945 & [0.920, 0.970] & 0.936 \\
    KV-cache quant.  & 0.988 & 0.986 & [0.974, 0.997] & 0.984 \\
    magnitude        & 0.990 & 0.990 & [0.955, 1.026] & 0.989 \\
    Gaussian noise   & 0.996 & 0.995 & [0.980, 1.011] & 0.997 \\
    GPTQ             & 0.996 & 0.996 & [0.988, 1.005] & 1.004 \\
    AWQ              & 1.003 & 1.003 & [0.991, 1.016] & 1.006 \\
    Wanda            & 1.005 & 1.007 & [0.970, 1.044] & 0.986 \\
    RTN              & 1.009 & 1.022 & [1.005, 1.039] & 1.014 \\
    SparseGPT        & 1.019 & 1.024 & [1.001, 1.048] & 1.009 \\
    \bottomrule
  \end{tabular}
  \end{fittable}
  \caption{Multiplicative deviation of each family's flip rate from its reference's fitted curve (802 configurations; intervals clustered by model). \emph{Held out} refits the curve without the family; \emph{overlap} restricts the family to $0.03\leq\KL\leq0.20$, where the families coexist, and fits the curve on the other families near that window.}
  \label{tab:families}
\end{table}

\begin{figure*}[!htb]
  \begin{minipage}[t]{0.48\textwidth}
    \centering
    \figorbox{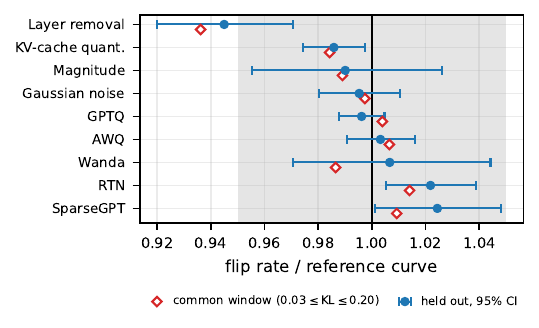}
    \caption{Deviation of each family's flip rate from its reference's fitted curve, with the curve refitted \emph{without} the family (blue, $95\%$ CI clustered by model; 802 configurations) and restricted to the window $0.03\leq\KL\leq0.20$ where all families coexist (red). The shaded band is $\pm5\%$. Numbers in Table~\ref{tab:families}.}
    \label{fig:families}
  \end{minipage}\hfill
  \begin{minipage}[t]{0.48\textwidth}
    \centering
    \includegraphics[width=0.85\linewidth]{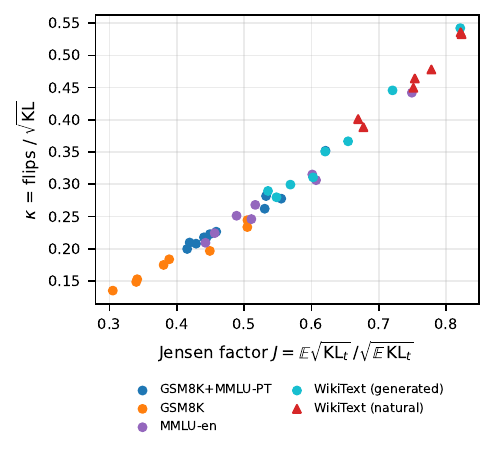}
    \caption{$r_{0.25}$ (Table~\ref{tab:mechanism}) against the Jensen factor $J$ of Eq.~\eqref{eq:jensen} for the 42 references with an fp32 output head; $J$ measures how evenly the divergence is spread across tokens, and it carries $70\%$ of the variance of $\log r_{0.25}$.}
    \label{fig:kappaj}
  \end{minipage}
\end{figure*}

\begin{figure*}[t]
  \centering
  \includegraphics[width=0.7\textwidth]{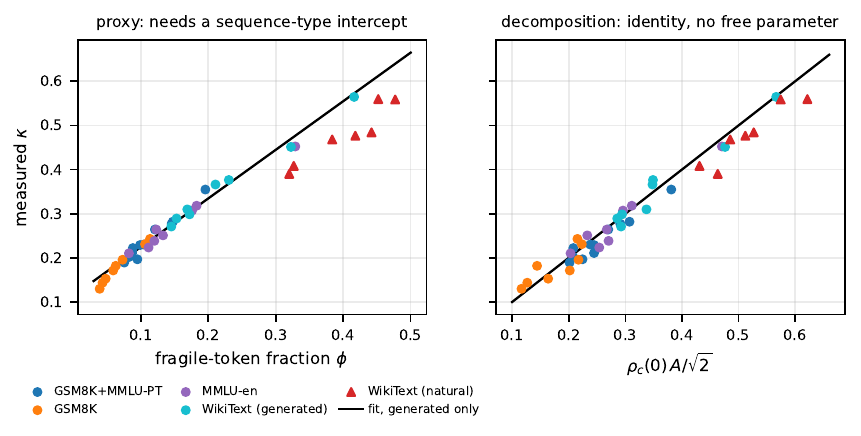}
  \caption{\emph{Left}: $\kappa$ against the fragile-token fraction $\phi$, with the line fitted on generated text; natural-text references (triangles) sit below it. \emph{Right}: the same references against $\rhoc A/\sqrt2$ with the identity line and no fitted parameter; the natural-text references move to within $7\%$ of the identity (median observed-to-predicted $0.93$), still slightly below it.}
  \label{fig:proxy-panel}
\end{figure*}

\begin{figure*}[!htb]
  \begin{minipage}[t]{0.48\textwidth}
    \centering
    \includegraphics[width=0.85\linewidth]{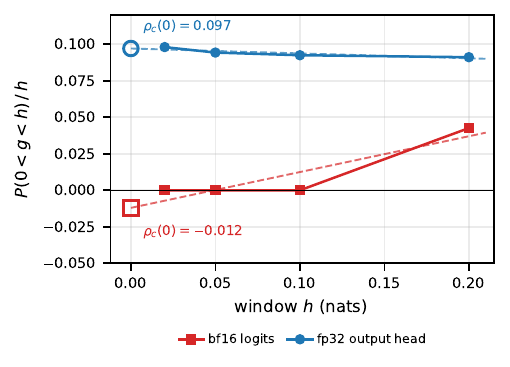}
    \caption{The estimate of $\rhoc$ on Gemma-3-4B with an fp32 output head (extrapolates to $0.097$) and on bf16 logits, where the grid leaves no margins below the first step and the extrapolation returns $-0.012$.}
    \label{fig:density}
  \end{minipage}\hfill
  \begin{minipage}[t]{0.48\textwidth}
    \centering
    \includegraphics[width=0.85\linewidth]{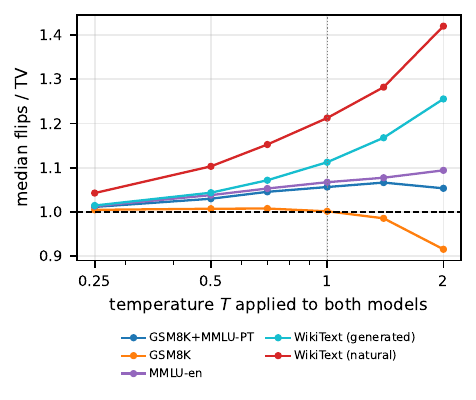}
    \caption{Median $\flips/\TV$ per corpus when both models' logits are divided by a temperature $T$, which leaves every flip unchanged; the ratio tends to one as $T\to0$ (42 fp32 references, top-64 tokens).}
    \label{fig:temperature}
  \end{minipage}

  \vspace{4ex}
  \centering
  \includegraphics[width=0.35\textwidth]{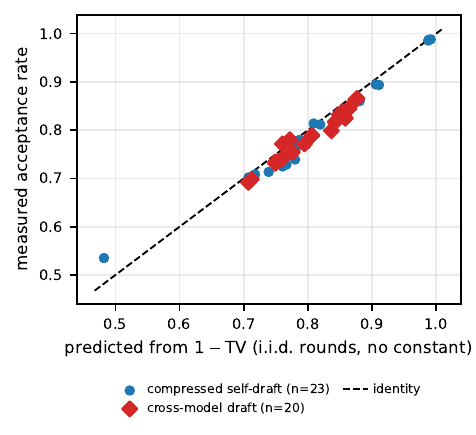}
  \caption{Greedy acceptance against the prediction from $1-\TV$ through independent rounds, no fitted constant (43 pairs). The round model overestimates acceptance for every pair ($+2.2$ and $+2.3$ points); total variation's slight overestimate of flips (flips$/\TV$ $0.97$--$0.99$) offsets part of it, hence its edge in stage 2 of Table~\ref{tab:spec}. Wall-clock speedup depends on kernels: a naive cost model ranks it at Spearman $+0.30$ and $+0.54$ on two targets.}
  \label{fig:spec}
\end{figure*}

\end{document}